\documentclass[a4paper,11pt]{article}
\usepackage{pos}
\usepackage{natbib}
\usepackage{cancel}
\usepackage{wrapfig}
\usepackage{subcaption}
\usepackage{mathtools}

\title{HMC with Normalizing Flows}

\author*[a]{Sam Foreman}
\author[b, c]{Taku Izubuchi}
\author[d]{Luchang Jin}
\author[a]{Xiao-Yong Jin}
\author[a]{James C. Osborn}
\author[b]{Akio Tomiya}

\affiliation[a]{Argonne National Laboratory,\\
  Lemont, IL 60439}

\affiliation[b]{RIKEN,\\
 2-1 Hirosawa, Wako, Saitama, 351-0198, Japan}

\affiliation[c]{Brookhaven National Laboratory,\\
 Upton, NY 11973}

\affiliation[d]{Dept. of Physics, University of Connecticut,\\
 Storrs, CT 06269}

\emailAdd{foremans@anl.gov}
\emailAdd{izubuchi@bnl.gov}
\emailAdd{luchang.jin@uconn.edu}
\emailAdd{xjin@anl.gov}
\emailAdd{osborn@alcf.anl.gov}

\abstract{%
    We propose using Normalizing Flows as a trainable kernel within the
    molecular dynamics update of Hamiltonian Monte Carlo (HMC).
    By learning (invertible) transformations that simplify our dynamics, we can
    outperform traditional methods at generating independent configurations.
    We show that, using a carefully constructed network architecture, our
    approach can be easily scaled to large lattice volumes with minimal
    retraining effort.
    The source code for our implementation is publicly available online at
    \href{https://www.github.com/nftqcd/fthmc}{github.com/nftqcd/fthmc}.
}

\FullConference{%
 The 38th International Symposium on Lattice Field Theory, LATTICE2021
  26th-30th July, 2021
  Zoom/Gather@Massachusetts Institute of Technology
}

\begin{document}
\maketitle
\section{\label{sec:intro}Introduction}
\subsection{%
    \label{subsec:gauge_theory}%
    2D \texorpdfstring{\(U(1)\)}{U(1)} Gauge Theory
}
\begin{wrapfigure}{r}{.33\columnwidth}
  \includegraphics[width=0.33\columnwidth]{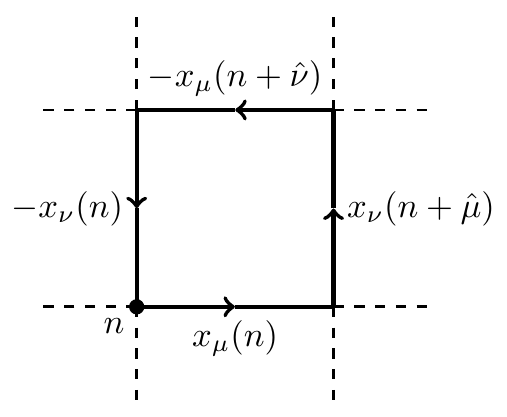}
  \caption{\label{fig:plaq} Plaquette \(x_{P}\).}
\end{wrapfigure}
Let \(U_{\mu}(n) = e^{i x_{\mu}(n)}\in U(1)\), with \(x_{\mu}(n)\in [-\pi,
\pi]\) denote the \emph{link variables}, where \(x_{\mu}(n)\) is a link at the
site \(n\) oriented in the direction \(\hat{\mu}\).
Our goal is to generate an ensemble of configurations, distributed according to
%
\begin{equation}
    p(x)\propto e^{-S(x)},\quad S(x) \equiv \sum_{P} 1 - \cos x_{P},
\end{equation}
where $S(x)$ is the Wilson action for the 2D \(U(1)\) gauge
theory\footnote{Explicitly, on a square lattice with periodic boundary
conditions.}, and \(x_{P} = x_{\mu}(n) + x_{\nu}(n+\hat{\mu}) -
x_{\mu}(n+\hat{\nu}) - x_{\nu}(n)\) is the sum of the links around the
elementary plaquette as shown in Figure~\ref{fig:plaq}.
For a given lattice configuration, we can define the topological charge as \(Q
= \frac{1}{2\pi}\sum_{P}\mathrm{arg}(x_{P}) \in\mathbb{Z}\), where
\(\mathrm{arg}(x_{P})\in [-\pi, \pi]\).
%

Traditional sampling techniques such as HMC are known to suffer from
\emph{critical slowing down}~\cite{Schaefer:2010hu}, a phenomenon characterized
by the freezing of the topological charge \(Q\) as we approach physical lattice
spacings.
This effect can be seen clearly in Figure~\ref{subfig:q8},~\ref{subfig:q16},
where \(Q\) typically remains stuck for the duration of the HMC trajectories.
In this work we describe a method for training a normalizing flow model that is
capable of sampling from different topological charge sectors, thereby reducing
the computational effort required to generate independent configurations.
%

\subsection{\label{subsec:ft}Field Transformations}
For a random variable \(z\) with a given distribution \(z \sim r(z)\), and an
invertible function \(x = f(z)\) with \(z = f^{-1}(x)\), we can use the change
of variables formula to write
\begin{equation}
    p(x) = r(z)\left|\det\frac{\partial z}{\partial x}\right| =
    r(f^{-1}(x))\left|\det\frac{\partial f^{-1}}{\partial x}\right|
\end{equation}
where \(r(z)\) is the (simple) prior density, and our goal is to generate
independent samples from the (difficult) target distribution \(p(x)\).
This can be done using \emph{normalizing flows}~\cite{rezende2015variational}
to construct a model density \(q(x)\) that approximates the target
distribution, i.e. \(q(\cdot)\simeq p(\cdot)\) for a suitably-chosen flow
\(f\).

\begin{figure}[htpb]
    \centering
    \includegraphics[width=\textwidth]{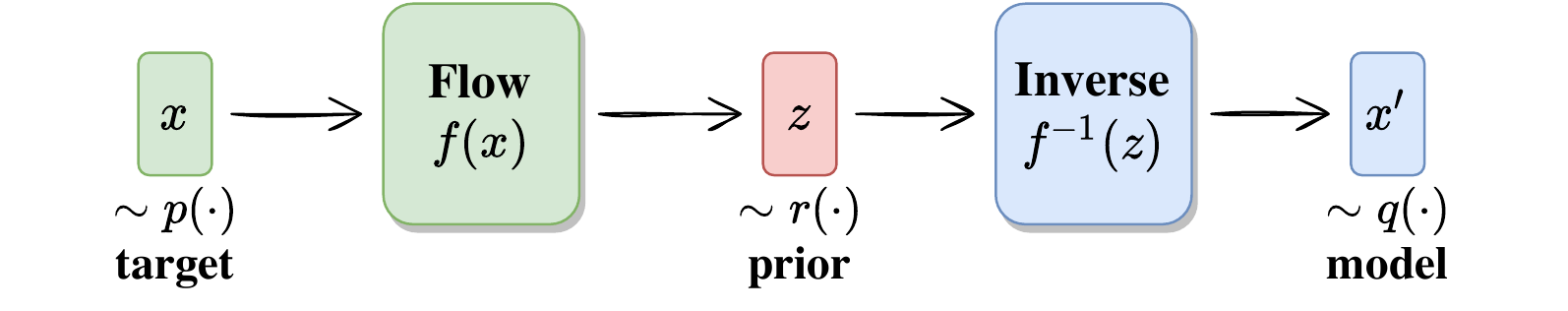}
    \caption{\label{fig:flow_model} Using a flow to generate data \(x'\). Image
    adapted from~\cite{weng2018flow}}
\end{figure}
We can construct a normalizing flow by composing multiple invertible functions
\(f_{i}\) so that \(x\equiv \left[f_{k}\circ f_{k-1}\circ \cdots \circ
f_{2}\circ f_{1}\right](z)\).
In practice, the functions \(f_{i}\) are usually implemented as \emph{coupling
layers}, which update an ``active'' subset of the variables, conditioned on the
complimentary ``frozen'' variables~\cite{Kanwar:2020xzo,Albergo:2021vyo}.
\subsection{\label{subsec:coupling_layers}Affine Coupling Layers}
A particularly useful template function for constructing our normalizing flows
is the affine coupling layer~\cite{DinhSB16,rezende2015variational},
\begin{align*}
    f(x_{1}, x_{2}) &= \left(e^{s(x_2)}x_{1} + t(x_{2}),\, x_{2}\right),
        \quad\text{with}\quad \log J(x) = \sum_{k}\left[s(x_{2})\right]_{k}\\
    f^{-1}(x'_{1}, x'_{2}) &= \left((x'_{1}-t(x'_{2}))e^{-s(x'_{2})},\, x'_{2}\right),
        \quad\text{with}\quad \log J(x') = \sum_{k}-\left[s(x'_{2})\right]_{k}
\end{align*}
where \(s(x_{2})\) and \(t(x_{2})\) are of the same dimensionality as \(x_{1}\)
and the functions act element-wise on the inputs.

In order to effectively draw samples from the correct target distribution
\(p(\cdot)\), our goal is to minimize the error introduced by approximating
\(q(\cdot)\simeq p(\cdot)\).
To do so, we use the (reverse) Kullback-Leibler (KL) divergence from
Eq.~\ref{eq:kl_div}, which is minimized when \(p=q\).
\begin{align}
    \label{eq:kl_div}
    D_{\mathrm{KL}}(q\|p) 
    &\equiv\int dy q(y)\left[\log q(y) - \log p(y)\right]\\
    &\simeq \frac{1}{N}\sum_{i=1}^{N} \left[\log q(y_{i})-\log p(y_{i})\right],
        \,\,\text{where}\,\, y_{i}\sim q
\end{align}
\section{\label{sec:trivializing_map}Trivializing Map}
Ultimately, our goal is to evaluate expectation values of the form
\begin{equation}
    \label{eq:exp_val}
    \langle \mathcal{O} \rangle = \tfrac{1}{\mathcal{Z}} \int dx\, \mathcal{O} (x) e^{-S(x)}.
\end{equation}
Using a normalizing flow, we can perform a change of variables \(x = f(z)\), so
Eq.~\ref{eq:exp_val} becomes
\begin{align}
    \langle \mathcal{O} \rangle 
    &= \frac{1}{\mathcal{Z}} \int dz \left|\det \left[ J(z) \right]\right|
        \mathcal{O} (f(z)) e^{-S(f(z))},
        \text{ where } J (z) = \frac{\partial f(z)}{\partial z} \\
    &= \frac{1}{\mathcal{Z}}\int dz \mathcal{O} (f(z)) e^{-S(f(z))
        + \log |\det[J(z)]|}.
\end{align}
We require the Jacobian matrix, \(J(z)\), to be:
\begin{enumerate}
    \item Injective (1-to-1) between domains of integration
    \item Continuously differentiable (\emph{or}, differentiable with
        continuous inverse)
\end{enumerate}
The function \(f\) is a \emph{trivializing map}~\cite{luscher2009} when
\(S(f(z)) - \log\left|\det J(z)\right| = \text{const.}\), and our expectation
value simplifies to
\begin{equation}
    \langle\mathcal{O}\rangle = 
    \frac{1}{\mathcal{Z}^{\ast}}\int dz\, \mathcal{O}(f(z)), \text{ where }
    \frac{1}{\mathcal{Z}^{\ast}} 
    = \frac{1}{\mathcal{Z}}\exp(-\text{const.}).
\end{equation}
\section{\label{sec:hmc_nf}Field Transformation HMC: \texttt{fthmc}}
\begin{figure}[htpb]
    \centering
    \includegraphics[width=\textwidth]{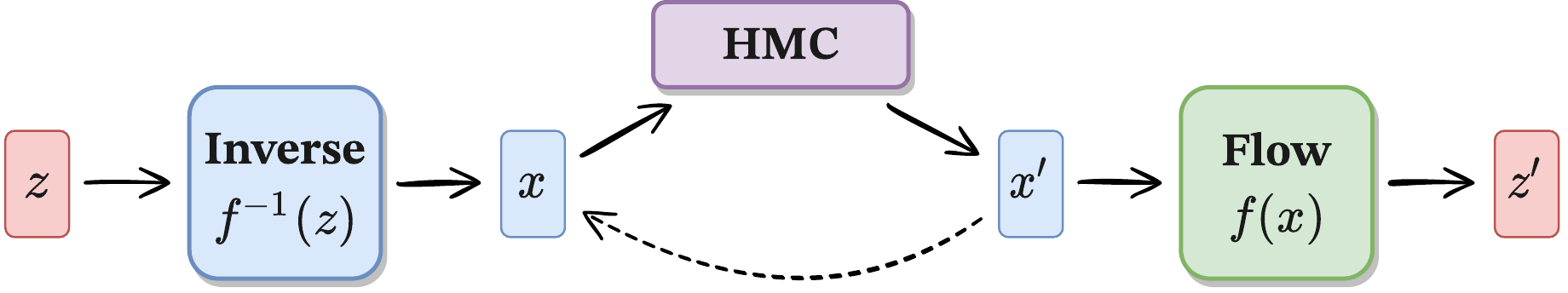}
    \caption{\label{fig:fthmc}Normalizing flow with inner HMC block.}
\end{figure}

We can implement the trivializing map defined in
Sec.~\ref{sec:trivializing_map} using a normalizing flow model.
For conjugate momenta \(\pi\), we can write the Hamiltonian as
\begin{equation}
    H(z, \pi) = \frac{1}{2}\pi^{2} + S(f(z)) - \log\left|\det J(f(z))\right|,
\end{equation}
and the associated equations of motion as
\begin{align}
    \dot{z} &= \frac{\partial H}{\partial \pi} = \pi \\
    \dot{\pi} &= -J(z) S'(f(z)) + \mathrm{tr}\left[ J^{-1}\frac{d}{dz} J \right].
\end{align}
If we introduce a change of variables, \(\pi = J(z)\rho = J(f^{-1}(x))\rho\)
and \(z = f^{-1}(x)\), the determinant of the Jacobian matrix reduces to \(1\),
and we obtain the modified Hamiltonian
\begin{equation}
    \tilde{H}(x, \rho) = \frac{1}{2}\rho^{\dagger}\rho + S(x) - \log|\det J|.
\end{equation}
As shown in Figure~\ref{fig:fthmc}, we can use a \emph{field transformation},
\(f^{-1}: z \rightarrow x\) to perform HMC updates on the transformed variables
\(x\), and \(f: x \rightarrow z\) to recover the physical target distribution.
\subsection{\label{subsec:hmc}Hamiltonian Monte Carlo (HMC)}
We describe the general procedure of the Hamiltonian Monte Carlo
algorithm~\cite{Betancourt:2017}.
\begin{enumerate}
    \item Introduce \(v \sim \mathcal{N} (0,\mathbb{I}_{n}) \in \mathbb{R}^{n}\)
        and write the joint distribution as
        \begin{equation}
            p(x, v) = p(x) p(v) \propto e^{-S(x)} e^{-\frac{1}{2} v^{T} v}
        \end{equation}
    \item Evolve the joint system \((\dot x, \dot v)\) according to
        Hamilton's equations along \(H=\text{const.}\) using the leapfrog
        integrator:
        \begin{equation}
            \textbf{ (a.)  } \tilde{v} \leftarrow v - \frac{\varepsilon}{2}\partial_{x}S(x)\quad
            \textbf{ (b.)  } x' \leftarrow x + \varepsilon \tilde{v}\quad
            \textbf{ (c.)  } v' \leftarrow \tilde{v} - \frac{\varepsilon}{2}\partial_{x} S(x')
        \end{equation}
    \item Accept or reject the proposal configuration using the
        Metropolis-Hastings test,
        \begin{equation}
            x_{i+1} = \begin{cases}
                x', \text{ with probability } 
                    A(x'|x) \equiv \min\left\{1, \frac{p(x')}{p(x)}%
                    \left|\frac{\partial x'}{\partial x^{T}}\right|\right\}\\
                x, \text{ with probability } 1 - A(x'|x)
            \end{cases}
        \end{equation}
\end{enumerate}
\subsection{\label{subsec:volume_scaling}Volume Scaling}
We use gauge equivariant coupling layers that act on plaquettes as the
base layer for our network architecture.
As in~\cite{Albergo:2021vyo}, these layers are composed of inner coupling
layers which are implemented as stacks of convolutional layers.
One advantage of using convolutional layers is that we can re-use the trained
weights when scaling up to larger lattice volumes.
Explicitly, when scaling up the lattice volume we can initialize the weights
of our new network with the previously trained values.
This approach has the advantage of requiring minimal retraining effort while
being able to efficiently generate models on large lattice volumes.
\section{\label{sec:results}Results}
\begin{figure}[htpb]
    \centering
    \begin{subfigure}[b]{\linewidth}
        \includegraphics[width=\linewidth]{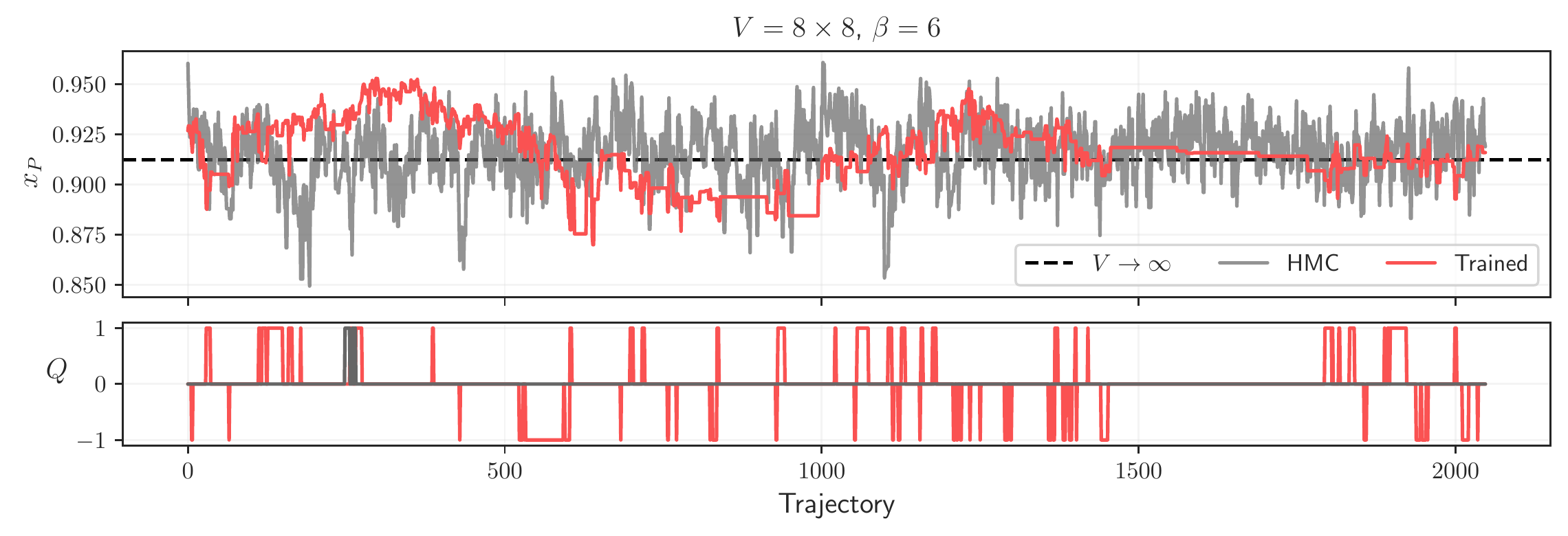}
        \caption{\label{subfig:q8}The average plaquette \(x_{P}\) and
            topological charge \(Q\) histories for the trained model and
        HMC at \(\beta = 6\) with \(V = 8 \times 8\).}
    \end{subfigure}
    \hfill
    \begin{subfigure}[b]{\linewidth}
        \includegraphics[width=\linewidth]{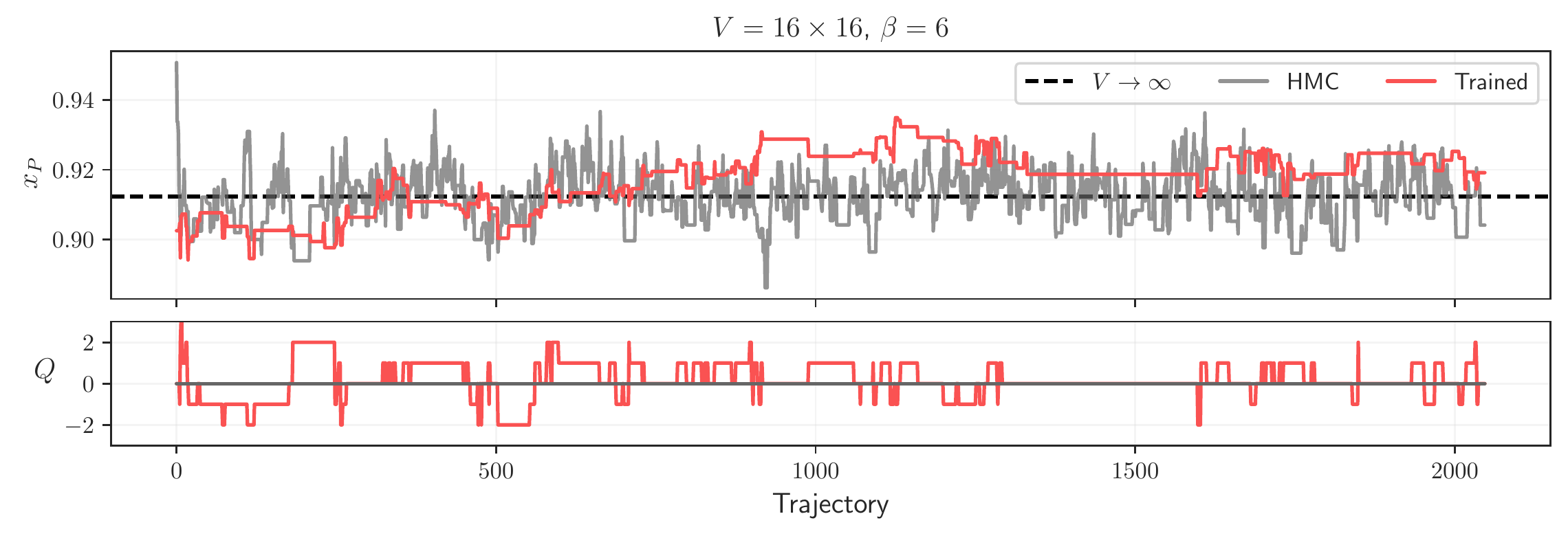}
        \caption{\label{subfig:q16}The same model from Figure~\ref{subfig:q8}
        used to generate configurations on \(V = 16\times16\) lattice.}
    \end{subfigure}
    \caption{\label{fig:histories}Comparison of lattice observables for both
        HMC and the trained model at \(V = 8\times 8\), and \(V =
    16\times16\).}
\end{figure}

For traditional HMC, we see in Figure~\ref{subfig:q8},\ref{subfig:q16}
that \(Q \simeq 0\) for across all trajectories for both \(8\times 8\) and
\(16\times 16\) lattice volumes.
Conversely, we see in Figure~\ref{subfig:q8},\ref{subfig:q16} that the trained
models are able to sample from multiple values of \(Q\) for both the \(8\times
8\) and \(16\times16\) volumes.

\begin{figure}
    \centering
    \includegraphics[width=\linewidth]{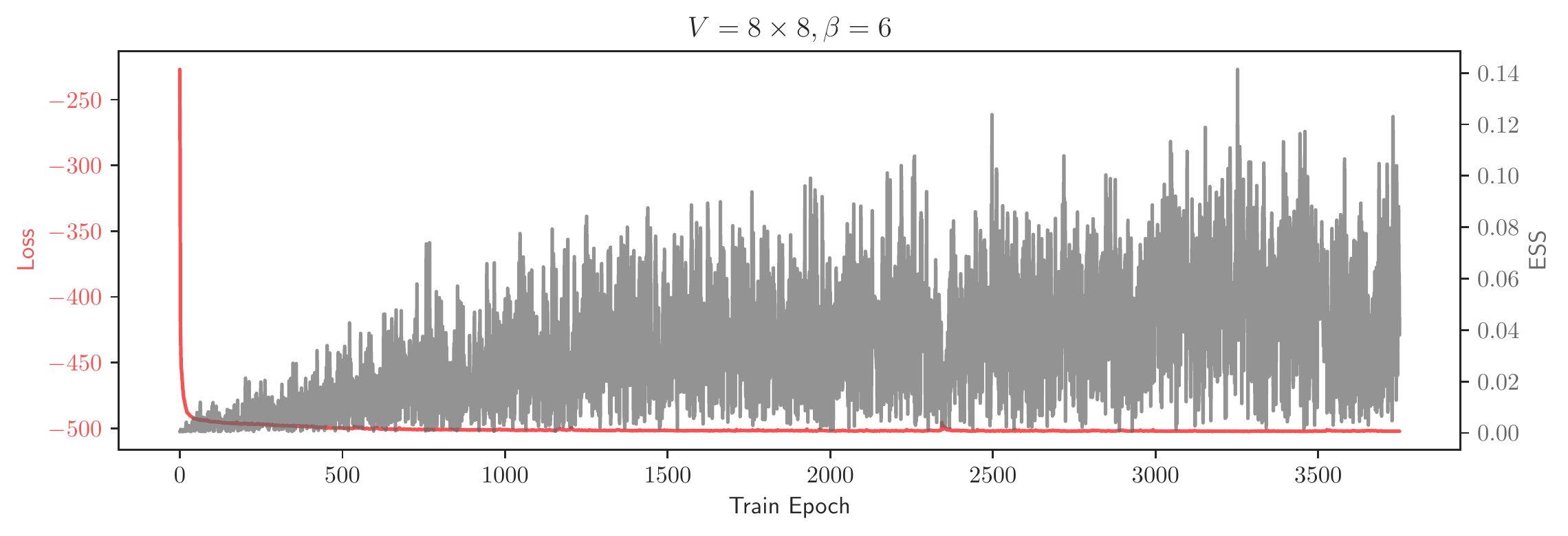}
    \caption{\label{subfig:loss}Loss and Effective Sample
        Size~\cite{2018arXiv180904129E} (ESS) vs train epoch at \(\beta = 6\)
    on \(V = 8\times 8\) lattice.}
\end{figure}

The results in Figure~\ref{subfig:loss} took \(\sim 4\) hours to train using a
single A100 Nvidia GPU.
The performance of the trained sampler is limited by the acceptance rate of the
proposed configurations, which in turn, is ultimately limited by the
computational resources used to train the model.
Because of this, we would expect a continued improvement in performance with
additional training.
For this relatively simple proof of concept, we were able to
demonstrate the usefuleness of our approach without requiring prohibitively
large upfront training costs.
\section{\label{sec:ack}Acknowledgments}
This research was supported by the Exascale Computing Project (17-SC-20-SC), a
collaborative effort of the U.S. Department of Energy Office of Science and the
National Nuclear Security Administration.
This research was performed using resources of the Argonne Leadership Computing
Facility (ALCF), which is a DOE Office of Science User Facility supported under
Contract DE\_AC02--06CH11357. 
This work describes objective technical results and analysis.
Any subjective views or opinions that might be expressed in the work do not
necessarily represent the views of the U.S. DOE or the United States
Government.
Results presented in this research were obtained using the Python
\citep{van1995python}, programming language and its many data science libraries
\cite{%
    matplotlib,
    harris2020array,
    ipython4160251%
}
%
\bibliographystyle{JHEP}
\bibliography{main}


\end{document}